# Ultra-Low-Power Spiking Neurons in 7 nm FinFET Technology: A Comparative Analysis of Leaky Integrate-and-Fire, Morris-Lecar, and Axon-Hillock Architectures

Logan Larsh,[1] Raiyan Siddique,[2] Sarah Sharif,[1] and Yaser Mike Banad[1,*]

[1]The School of Electrical and Computer Engineering, The University of Oklahoma, Norman, OK, 73019, United States
[2]Department of Electrical and Computer Engineering, Olin College of Engineering, Needham, MA, 02492, United States
*bana@ou.edu

## ABSTRACT

Neuromorphic computing aims to replicate the brain's remarkable energy efficiency and parallel processing capabilities for large-scale artificial intelligence applications. In this work, we present a comprehensive comparative study of three spiking neuron circuit architectures—Leaky-Integrate-and-Fire (LIF), Morris–Lecar (ML), and Axon-Hillock (AH)—implemented in a 7 nm FinFET technology. Through extensive SPICE simulations, we explore the optimization of spiking frequency, energy per spike, and static power consumption. Our results show that the AH design achieves the highest throughput, demonstrating multi-gigahertz firing rates (up to 3 GHz) with attojoule energy costs. By contrast, the ML architecture excels in subthreshold to near-threshold regimes, offering robust low-power operation (as low as 0.385 aJ/spike) and biological bursting behavior. Although LIF benefits from a decoupled current mirror for high-frequency operation, it exhibits slightly higher static leakage compared to ML and AH at elevated supply voltages. Comparisons with previous node implementations (22 nm planar, 28 nm) reveal that 7 nm FinFETs can drastically boost energy efficiency and speed—albeit at the cost of increased subthreshold leakage in deep subthreshold regions. By quantifying design trade-offs for each neuron architecture, our work provides a roadmap for optimizing spiking neuron circuits in advanced nanoscale technologies to deliver neuromorphic hardware capable of both ultra-low-power operation and high computational throughput.

## Introduction

As artificial intelligence (AI) scales to address pressing global challenges—ranging from disease eradication to energy scarcity and world hunger—its computational demands grow ever more complex. Ironically, one of the foremost barriers to solving these problems is the immense power consumption of conventional hardware, which is pushing thermal densities to critical limits and driving energy costs upward even as performance gains plateau. Neuromorphic computing, inspired by the brain's highly efficient spike-based signaling, offers a promising path to handle large-scale problems under stringent energy constraints[1–5]. The human brain, for instance, consumes approximately 72 kJ (20 Wh) of energy per day, whereas a typical central processor requires around 8640 kJ (2400 Wh) per day, achieving only a fraction of the brain's computational capability[6,7]. In light of these challenges, this paper explores ultra-low-power spiking neuron architectures implemented in a 7 nm FinFET technology node—demonstrating significantly reduced energy per spike and paving the way for more sustainable, high-performance AI systems[8].

Neuromorphic architectures such as IBM's TrueNorth and Intel's Loihi have demonstrated how biologically inspired designs can drastically reduce power consumption in large-scale AI systems[9–11]. However, these implementations generally rely on older technology nodes, where fundamental limits on transistor density, leakage control, and on-chip interconnects constrain further improvements in performance and efficiency. Consequently, pushing neuromorphic computing to the next level requires harnessing advanced process nodes—particularly 7 nm FinFET technology—to unlock lower spiking energy and higher operational frequencies. By leveraging this cutting-edge device technology, our work aims to systematically optimize circuit-level parameters (e.g., supply voltage, capacitances) to achieve sub-picojoule spiking energies while preserving the essential spike-based information processing that makes neuromorphic systems so compelling.

While advanced process nodes can drastically reduce transistor-level power consumption, the heart of neuromorphic computing still lies in emulating the spiking mechanism of real neurons. At the core of this biologically inspired design is the single neuron, which consists of three primary components: the dendrites, the cell membrane (soma), and the axon (Figure 1a)[7,12]. The dendrites receive electrochemical signals from preceding neurons, which are then integrated by the cell membrane. Once a critical threshold is reached, an action potential is generated and propagated through the axon to other neurons. This spike-based information processing occurs across billions of interconnected neurons in the brain, allowing it to efficiently handle complex tasks with remarkable energy efficiency (Figure 1(b)-1(d)). Notably, the brain's energy efficiency stems from the dynamic charge integration and reset mechanisms governing neuronal firing. By mimicking these biological properties with CMOS circuits, researchers have paved a pathway toward more energy-efficient artificial neuron designs.

Building on these fundamental principles, significant gains in neuromorphic performance have been achieved by moving to scaled CMOS technology nodes. For example, researchers have demonstrated significant gains by transitioning from older CMOS technology nodes toward smaller geometries. For instance, Indiveri et al.[13] introduced a CMOS integrate-and-fire (IF) neuron circuit that consumed only 1.5 µW of power, albeit with a limited spiking frequency and was implemented in a 1.5 µm technology node common to early neuromorphic chips. In the last decade, Wu et al.[14] implemented a LIF neuron in 180 nm CMOS, achieving an energy efficiency of 9.3 pJ per spike and updating synaptic weights to highlight the adaptability of spiking neuron circuits[15,16]. More recently, Vuppunuthala et al.[17] showcased a 65 nm LIF neuron with sub-threshold source-coupled logic (STSCL), further reducing spiking energy to 3.6 pJ per spike. Kyu-Bong Choi et al. [18] introduced an IF neuron circuit based on a Split-Gate Positive Feedback Device combined with CMOS at the 28 nm node, reducing energy consumption to 0.25 pJ per spike. Similarly, Chatterjee et al.[19] demonstrated FinFET-based neuron circuits consuming just 6.3 fJ per spike at 2 MHz in 14 nm FinFET technology. Although other research efforts explore modified FET structures[20–22] or alternative device technologies[23,24]—often achieving promising energy efficiencies and unique operational modes—widespread industry adoption remains centered on conventional CMOS, thanks to its maturity and established fabrication infrastructure.

In this study, we evaluate three neuron circuit architectures in a 7 nm FinFET technology,[25,26] optimizing their parameters for a refined balance between high-frequency performance and energy efficiency. The first architecture (Fig. 1(e)) is the Leaky-Integrate-and-Fire neuron[27], as the most common circuit abstraction, featuring a controlled leak and threshold-based spiking.[28–30] Incoming synaptic currents gradually charge the membrane potential until it crosses a predefined threshold, prompting an action potential and subsequent reset[31]. The "leaky" aspect mimics biological neurons by allowing the membrane voltage to decay over time, while the reset mechanism parallels the refractory period, preventing immediate successive firings[32]. The second neuron circuit architecture (Fig. 1f) is the Morris-Lecar (ML) neuron[33], a simplified biophysical model that directly drives the membrane and includes the reset mechanism to mimic the ionic currents behind depolarization and repolarization. By adjusting transistor operating points, it can produce a broad range of spiking behaviors, demonstrating how small parameter changes yield distinct firing regimes. The third architecture (Fig. 1g) is the Axon-Hillock (AH) neuron, which implements a hybridized membrane/reset mechanism reminiscent of the axon-hillock region in biological neurons[34]. The design effectively emulates the triggering of an action potential at the axon-hillock, followed by a rapid return to the resting states, preventing continuous spiking. Balancing threshold-based simplicity with the richer dynamics of biophysical models, the AH neuron is a versatile option for compact neuromorphic hardware.

Despite differences in these circuit architectures, all three models operate on the fundamental principle of integrating synaptic currents until reaching a critical threshold, triggering an output spike, and resetting the stored energy. However, comparing their performance and efficiency at the same scaled technology node and their best achievable performance is not straightforward, given that each design emphasizes different aspects of neuronal behavior and utilizes different CMOS technology nodes. Indeed, tuning parameters can significantly alter the firing characteristics and energy consumption profiles. As a result, a direct cross-comparison of these architectures in the same 7 nm FinFET process node requires careful calibration of design parameters, along with standardized metrics that capture both spike-generation dynamics and overall power efficiency. Notably, the original demonstrations of these LIF, ML, and AH neuron circuits generated spikes at approximately 1 kHz with a spiking energy range of ~1 fJ[27,33,34]. To further quantify and optimize these improvements, extensive parametric sweeps were performed on key design parameters, including membrane and reset capacitances, and the supply voltage. Consequently, we identify the optimal trade-offs between spiking frequency and energy consumption across these configurations. Specifically, we examined how variations in capacitance values influenced charge integration efficiency and reset dynamics, revealing critical dependencies between circuit parameters and performance. The insights gained from these simulations provided a comprehensive characterization of artificial neuron behavior at nanoscale dimensions and supported the feasibility of energy-efficient neuromorphic computing in advanced semiconductor technologies.

In the following sections, we delve into each of the three neuron models—LIF, ML, and AH—and present detailed simulation results supporting our central hypothesis. By systematically optimizing circuit-level parameters within a 7 nm FinFET process, we demonstrate how these designs can be pushed to regimes of exceptional efficiency and performance. Our findings emphasize the tremendous potential of cutting-edge CMOS processes in achieving sub-picojoule spiking energies and gigahertz-level firing rates—thereby paving the way for large-scale, energy-efficient neuromorphic systems.

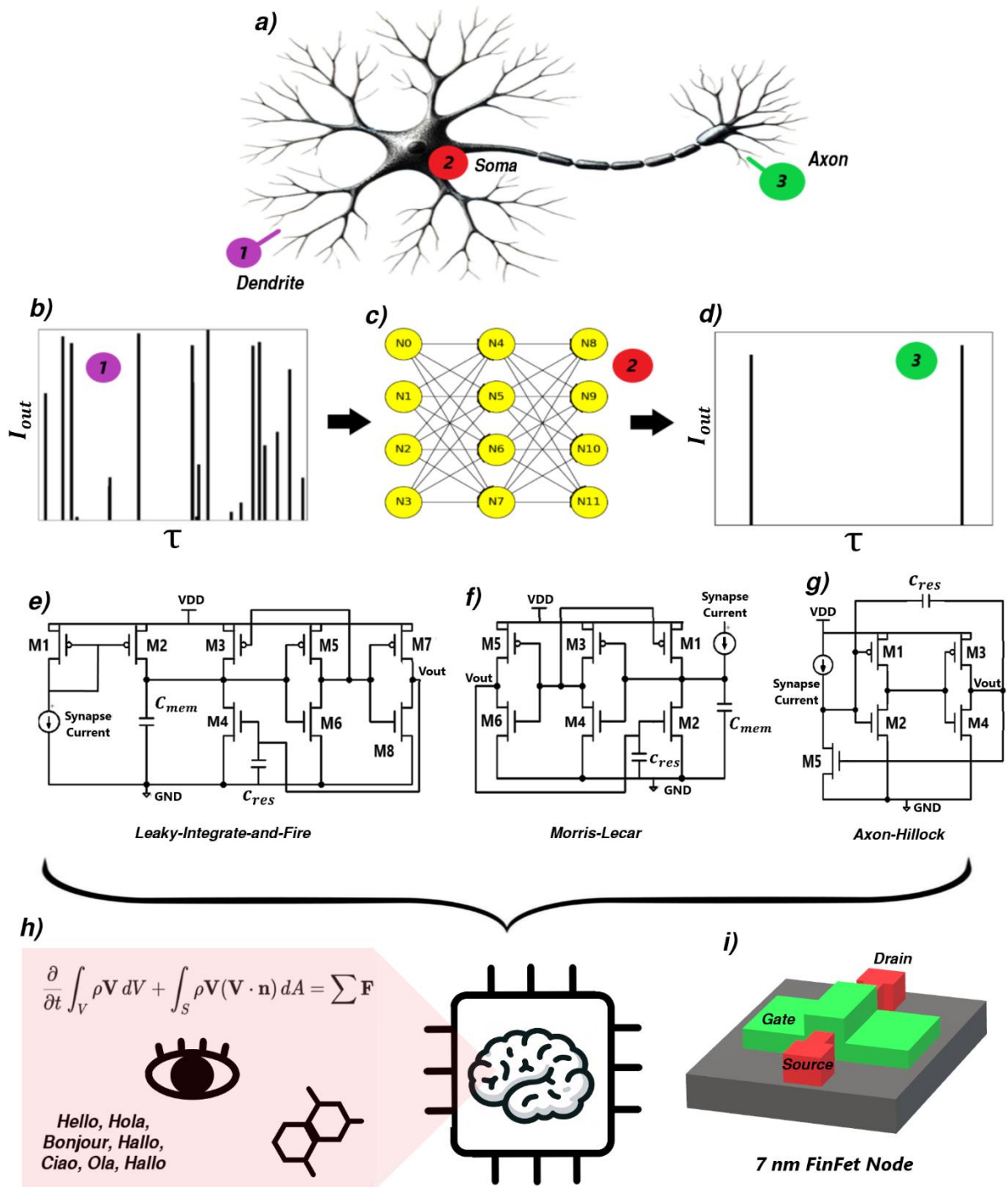

Figure 1. (**a**) Biological neurons consist of an axon, dendrites, and a soma. The dendrites experience a stochastic synapse input (**b**) from preceding neurons in the network. The soma, also referred to as the cellular membrane, integrates the incoming synaptic current to generate spikes (**d**) among other neurons in the network (**c**). (**e**) The Leaky Integrate and Fire (*LIF*) neuron implements a current mirror (M1-2), a membrane mechanism (M3-4), and a reset mechanism (M5-8), presented by Besrour et al.[27] (**f**) The Morris-Lecar (ML) neuron implements a directly driven membrane (M1-2), and reset mechanism (M3-6), presented by Sourikopolous et al.[33] (**g**) The Axon-Hillock (AH) neuron implements a hybridized membrane/reset mechanism, presented by Danneville et al.[34] This is accomplished by holding the integrated synapse current (M5) and resetting at the spiking threshold (M1-4). (**h**) The sensory signals and integral momentum balance illustrate how fundamental physics and sensory data streams can inspire neuromorphic design. (**i**) All models are simulated using a 7 nm FinFET technology node to demonstrate leveraging superior electrostatic control to achieve high efficiency and enhanced spiking frequency. Large-scale integration of 7 nm FinFET-based neurons pave the way for next-generation artificial intelligence hardware designs [1,25,26].

## Results

### Neuron Model Equations and Circuit Implementation

This section elaborates on the model equations and the circuit designs of LIF, ML, and AH neurons, demonstrating their alignments with biological neuron behavior[27,33,34]. Figure 2 shows the circuit simulation results of the three neurons for $V_{supp}$ = 0.2 V, $I_{syn}$ = 100 nA, and the same capacitor sizing of $C_{mem} = C_{rest}$ = 1fF. The dynamics of each neuron model can be understood through their governing equations, which directly relate to their circuit implementations.

### Leaky-Integrate-and-Fire Model

In a classical LIF neuron, the membrane potential $V_m$ evolves according to a first-order differential equation capturing both the integration of synaptic input and the gradual 'leak' of charge:

$$\tau_m \frac{dV_m}{dt} = -[V_m(t) - V_{rest}] + RI_{in}(t), \quad Eq.(1)$$

where $\tau_m$ represents the membrane time constant, $V_{rest}$ is the resting potential of the neuron, and $R$ is the effective membrane resistance. The net input current $I_{in}(t)$ is the sum of all synaptic currents and any external bias sources. In the LIF circuit (Figure 1(e)), the membrane capacitor (labeled $C_{mem}$) represents the "charge storage" element that integrates synaptic input over time, and the leak resistor (implemented by transistors acting in subthreshold) models the continuous decay of the membrane potential toward $V_{rest}$. The current mirror (transistors M1–M2) sources the synaptic current onto $C_{mem}$, matching the right-hand side of Equation (1), $RI_{in}(t)$.[35,36] When a presynaptic spike arrives, the synaptic current raises $V_m$ toward the threshold. This can be expressed using a simplified exponential form:

$$I_{syn}(t) = I_0 e^{-\frac{t}{\tau_{syn}}}, \quad Eq.(2)$$

which corresponds to charging $C_{mem}$ through the current mirror. Meanwhile, M3–M4 establish the membrane's "leaky" behavior by providing a controlled discharge path toward the resting potential, thereby emulating the passive decay in a biological neuron:

$$I_{leak} = \frac{V_m - V_{rest}}{R}, \quad Eq.(3)$$

Once the membrane voltage $V_m$ crosses a predefined spiking threshold, the reset mechanism (via M5–M8 transistors) quickly discharges $C_{mem}$, returning the membrane node to baseline. This reset action mimics the rapid downstroke of an action potential and prevents immediate successive firings, akin to the refractory period in real neurons[7,32]. Figure 2(a) shows the membrane capacitor voltage (middle waveform) rising from its resting level as it integrates the input current (top waveform). When the threshold is reached, the reset circuitry triggers a brief, sharp output spike (bottom waveform) and quickly brings $V_m$ back to baseline. Figure 2(a) illustrates repeated spiking over time, highlighting the rhythmic nature of the LIF's integrate-and-fire cycle. This rapid and repeatable spiking pattern in the tens-of-nanoseconds range suggests a similarity to cortical neurons, which produce a quasi-periodic rhythmic firing [7,12,32]. The arrangement of components in this design preserves high-frequency operation by preventing coupling with the preceding synapse current but at the cost of higher energy overhead.

### Morris–Lecar Model

Originally devised to capture excitable membrane dynamics in biological cells, the Morris–Lecar (ML) model focuses on two principal ionic currents—calcium and potassium—that respectively depolarize and repolarize a neuron's membrane potential. In mathematical form, the membrane voltage $V_m$ evolves according to[37–39]:

$$C\frac{dV_m}{dt} = I_{ext} - g_{Ca} m_\infty(V_m)(V_m - E_{Ca}) + g_k W(V_m - E_k) + g_L(V_m - E_L), \quad Eq.(4)$$

where $C$ is the membrane capacitance, $I_{ext}$ is the net external or bias current, and $g_{Ca}, g_k, g_L$ are conductances for the calcium, potassium, and leak pathways, respectively. $E_{Ca}, E_k, E_L$ are their respective reversal potentials, $W$ is a gating variable controlling the potassium-channel activation, and $m_\infty(V_m)$ describes the fraction of open calcium channels at a given voltage $V_m$. To include bursting —groups of rapid spikes separated by quiescent phases—researchers often add a slower ionic term or modify the gating functions, sometimes referred to as a "square-wave burster" extension[39]:

$$I_{burst} = a\Phi(V_m, \ldots), \quad Eq.(5)$$

where $a$ scales the bursting contribution, and $\Phi(.)$ encodes the slower dynamics that produce clusters of spikes.

In the ML neuron circuit (Figure 1(f)), the design replicates these biophysical behaviors using transistors. M1–M2 feed the incoming synapse current directly into the membrane capacitor, removing the decoupling mirror found in the LIF neuron. This approach makes the membrane voltage more sensitive to changes in synaptic input, akin to the rapid calcium influx in biological neurons. M3–M6 form adjustable conduction paths that correspond to potassium and leak currents. By tuning these devices' operating points (e.g., widths, lengths, and bias voltages), the circuit can emulate the slower gating dynamics in the original Morris–Lecar equations, including the delayed repolarization needed for repetitive spiking. This "no-decoupling" design reduces resistive overhead and allows the transistor network to mimic the voltage-dependent influx of calcium channels: $m_\infty(V_m)$. Transistor operating conditions in various regions (near-threshold to saturation) serve as the adjustable conductances $g_k$ and $g_L$. By tuning bias voltages and transistor widths/lengths, we replicate the gating effects seen in the original Morris–Lecar equations, including slow potassium channel activation ($W$) and a baseline leak current ($V_m - E_k$). When the circuit is designed with an auxiliary "slow current" branch or capacitive feedback path, we can implement the additional term $I_{burst}$ in Eq. (5). As a result, the membrane voltage periodically oscillates above the firing threshold, generating clusters of spikes like thalamocortical bursts in biological neurons[39,40].

Recordings of this neuron's output, captured across various supply voltages and current levels, reveal a capacity for both continuous periodic firing and bursts of spikes (Figure 2(c), 2(d), bottom waveform). Under bursting conditions, a transient cluster of spikes emerges, followed by a pause before the next burst sequence, all while consuming less energy per individual spike compared to the first circuit. By driving the membrane capacitor directly and employing transistor sub-circuits to represent the voltage-dependent calcium and potassium channels, the Morris–Lecar neuron captures a broader range of excitability patterns than the more abstract LIF model. This added complexity enables both continuous periodic firing and bursting modes, key properties of real neurons that can be harnessed for advanced neuromorphic computations—all while maintaining relatively low energy consumption compared to less biologically aligned designs.

**Axon-Hillock Model**

Unlike the Leaky-Integrate-and-Fire (LIF) and Morris–Lecar (ML) neurons, the Axon-Hillock (AH) neuron (Figure 2(b)) merges the membrane and reset nodes into a single capacitor[34]. This approach achieves spike generation and membrane discharge with fewer transistors, reducing both static leakage paths and dynamic power overhead. Conceptually, the model can be viewed as a streamlined LIF framework, often approximated by:[41]

$$C\frac{dV_m}{dt} = I_{syn}(t) - I_{leak}(V_m) \quad Eq.(6)$$

Here, $C$ is the sole capacitor in the circuit, fulfilling both "membrane" and "reset" roles. $I_{syn}(t)$ is the synaptic (input) current charging the capacitor. $I_{leak}(V_m)$ represents the collective leakage pathways that slowly pull $V_m$ back toward its resting level. In the transistor-level implementation (Figure 1(g)), M5 handles the incoming synapse current, charging the single capacitor $C_{res}$. Meanwhile, M1–M4 detect when the capacitor's voltage crosses a threshold, triggering a fast reset discharge back to baseline. This closely resembles how the biological axon hillock region triggers an action potential: once a neuron's membrane potential surpasses a critical value, a rapid spike ensues, followed by a refractory-like period in which the voltage is reset.

Due to its minimal design and fewer electrical paths to dissipate charge, the AH neuron exhibits exceptionally small energy per spike, particularly in the subthreshold and near-threshold voltage range. Under moderate input current, the neuron's membrane voltage smoothly ramps until a critical threshold, then instantly discharges—yielding stable, quasi-periodic spikes (Figure 2(b), bottom waveform) analogous to cortical neuron firing patterns[7,39]. Because one capacitor serves both the integration and reset functions, minor current backflow to the input side can occur during a spike's onset. However, the overall energy penalty remains minimal compared to designs with separate membrane and reset capacitors. These observations underscore how the AH circuit's single-capacitor design not only simplifies hardware implementation but also offers a compelling combination of high spiking frequency, low energy per spike, and robust operation at nanoscale process nodes. As with the LIF and ML neurons, fine-tuning device dimensions, bias currents, and operating voltages can further optimize the AH neuron for targeted performance metrics—ranging from ultra-low-power neuromorphic systems to high-speed signal processing in advanced AI applications.

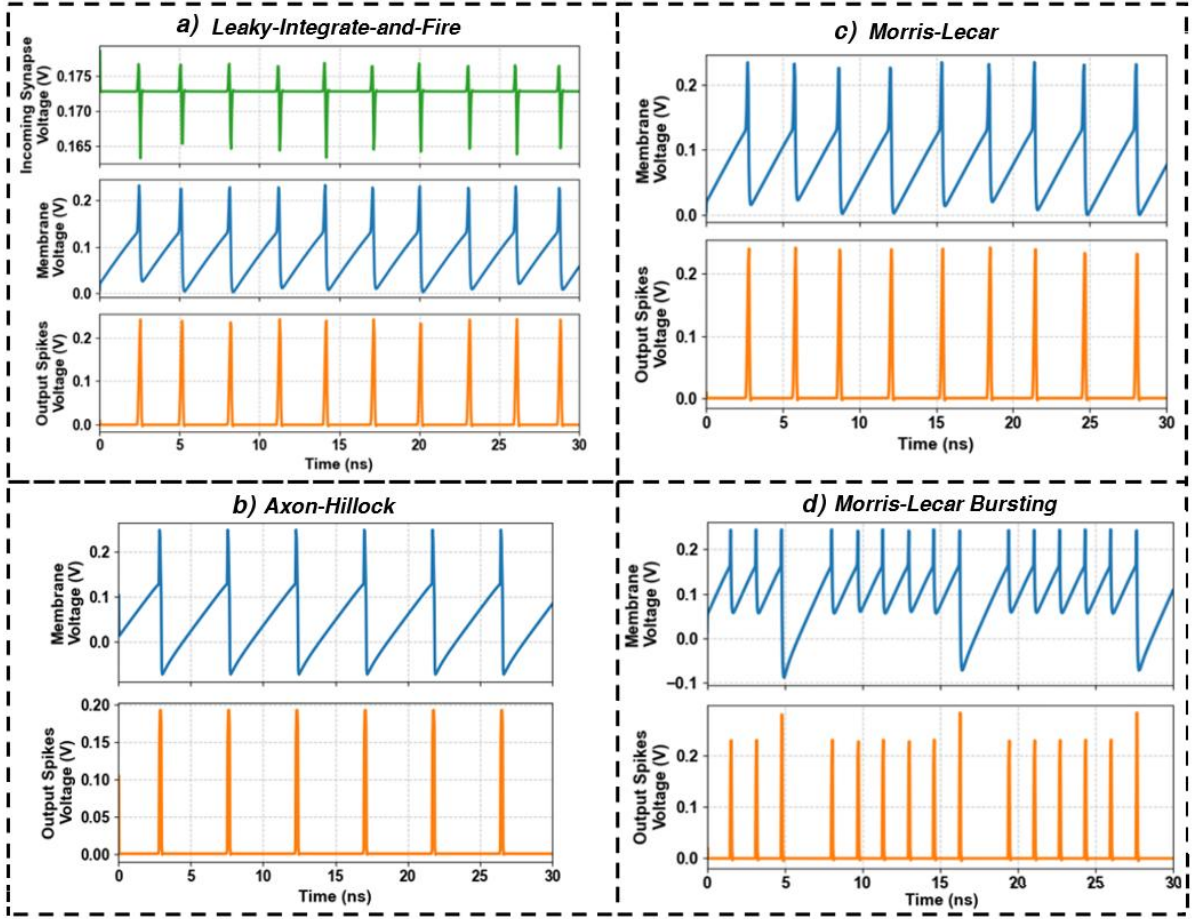

Figure 2. The behavioral waveforms of all neurons are demonstrated at sub-threshold supply voltages with a constant 100 nA synapse current and 1fF-1fF capacitor sizes. **(a, top)** The incoming synapse voltage is relatively stable due to the LIF current mirror. **(a, middle)** The membrane voltage then integrates the experienced synaptic current before triggering a spike upon the reset capacitor **(a, bottom)**. **(b, top)** The directly driven reset capacitor of the AH neuron integrates over time, triggering an output spike **(b, middle)** by which pulls itself down to 0. **(c, d, top)** The ML neuron performs with dual modality, generating both periodic and bursting spike patterns relative to the excitation caused by the synapse current vs. supply voltage. The resulting spikes of SML behaviors are represented in **(c, d, bottom)**

**Parametric Analysis of Performance–Efficiency Trade-Offs**

To systematically explore the trade-offs between energy per spike ($E_{spike}$) and spiking frequency ($f_{spike}$), each neuron design underwent a parametric sweep over supply voltages and capacitor sizes. Specifically, we varied the supply voltage from **0.1 V** to **0.9 V** in **0.05 V** increments, while the membrane $C_{mem}$ and reset $C_{res}$ capacitance took on values from **0.1 fF to 0.9 fF** in **0.1 fF** steps, and then integer steps from **1 fF** to **10 fF**. This range of capacitances spans very small values often desirable in neuromorphic research, as well as larger values aligned with older process nodes (e.g., 180 nm, 65 nm) for a direct benchmark comparison. By capturing spike energy and spike frequency for each combination of $C_{mem}$, $C_{res}$, and supply voltage, we assessed how quickly each neuron design charges toward the threshold and discharges upon spiking. In practice, minimizing $E_{spike}$ while maximizing $f_{spike}$ reflects an ideal balance between power efficiency and computational throughput. To enable meaningful comparison across a large design space, both $f_{spike}$ and $E_{spike}$ were normalized to their global minimum and maximum values observed in the sweeps. As shown in Equations (7)–(9), a negative sign is included in the energy term so that lower $E_{spike}$ maps to a higher normalized output. This ensures that improvements in efficiency shift the aggregated optimization score in the same positive direction as improvements in frequency. Configurations nearing an overall score of **1** thus represent high-frequency, low-energy solutions, commonly found in the deep-subthreshold **or** near-threshold domains, where device currents remain modest yet switching remains effective.

$$E_{spike_{norm}} = -1\left(\frac{E - E_{min}}{E_{max} - E_{min}}\right) \quad \text{Eq. (7)}$$

$$f_{spike_{norm}} = \frac{f - f_{min}}{f_{max} - f_{min}} \quad \text{Eq. (8)}$$

$$\text{Optimization Score} = E_{spike_{norm}} f_{spike_{norm}} \quad \text{Eq. (9)}$$

In the sections that follow, we detail how each of the three neuron circuits—BLIF, SML, and DAH—respond to these parameter sweeps, highlighting the configurations that yield the best balance of spiking speed and energy consumption within a 7 nm FinFET process.

**LIF Neuron Optimization**

Figure 3 illustrates how variations in membrane and reset capacitances affect the LIF neuron's spiking energy (EPS), firing rate, and overall optimization score relative to the average performance measured across the full supply voltage range. Figure 3(a) shows that EPS drops markedly for smaller reset capacitances ($C_{res}$ < 2.5 fF) because less stored charge is dissipated during each reset event. Indeed, the lowest measured EPS of 1.26 aJ occurs at $C_{mem}$ = 0.79 fF and $C_{res}$ = 0.1 fF (purple marker), underscoring how reset capacitance directly impacts energy consumption in this model. Meanwhile, spiking frequency depends more strongly on $C_{mem}$: as shown in Figure 3(b), reducing $C_{mem}$ accelerates the charging rate toward the threshold, yielding a peak frequency of 933 MHz at $C_{mem}$ = 0.5 fF and $C_{res}$ = 0.1 fF. Balancing these two competing goals leads to an optimal region in Figure 3(c) ($C_{mem}$ ≈ 0.69 fF, $C_{res}$ ≈ 0.2 fF), where the neuron achieves both moderate EPS (~2 aJ) and a high spiking frequency (~0.75 GHz).

Subsequent supply-voltage sweeps (Figure 3d–f) highlight how the LIF circuit responds to changes in $V_{supp}$. Although the energy per spike increases nearly monotonically from 0.52 aJ to 31.34 aJ as $V_{supp}$ rises from 0.1 V to 0.7 V, the spiking frequency reaches a maximum of 1.66 GHz at 2.35 aJ EPS near $V_{supp}$ = 0.38 V—a direct manifestation of the LIF equation's interplay between leakage, charging current, and threshold reset. Thus, in alignment with the analytical LIF framework introduced earlier, carefully sizing the membrane and reset capacitors while tuning the supply voltage can yield spikes within the range of atto Joule energy consumption and giga Hertz frequencies.

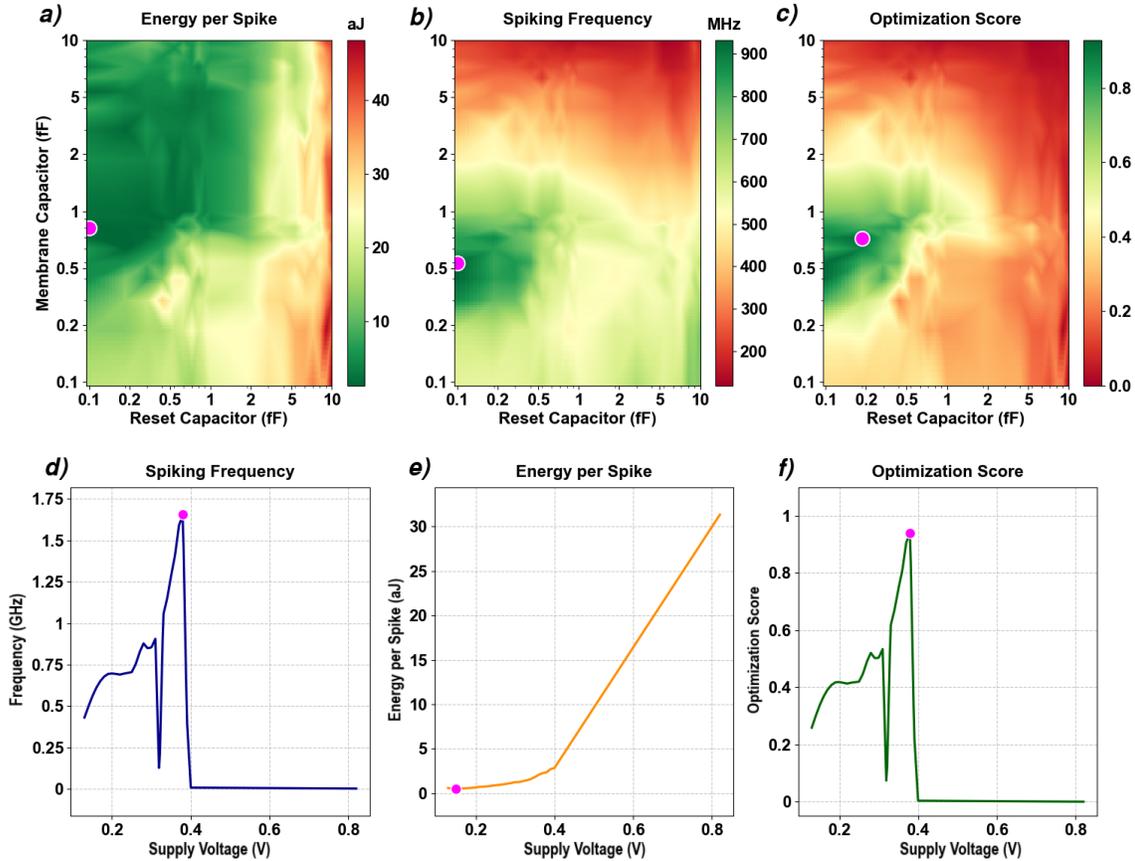

Figure 3. Parametric analysis of the Leaky-Integrate-and-Fire (LIF) neuron in 7 nm FinFET technology. (a–c) Contour plots showing the influence of the membrane capacitance (vertical axis) and reset capacitance (horizontal axis) on (a) Energy per spike (aJ), (b) Spiking frequency (MHz), and (c) An overall optimization score that balances high frequency with low energy. For a fair comparison, the average values across the supply voltage ranges (0.1V-0.7V) are used and the pink markers highlight regions of interest, including the minimum observed energy (~1.26 aJ) and maximum observed frequency (~933 MHz). (d–f) Line plots of the LIF neuron's performance under a supply-voltage sweep, using the best-performing capacitor pair identified in (a–c). (d) Spiking frequency peaks at 1.66 GHz

around 0.38 V,(e) Energy per spike ranges from below 1 aJ in deep subthreshold to tens of aJ above 0.6 V, and (f) The optimization score indicates an optimal operating point near 0.38 V, balancing both high frequency and low energy per spike.

**ML Neuron Optimization**

The ML neuron demonstrates a similar capacitance relationship to the LIF neuron. Figure 4(a–b) demonstrates that reducing the membrane capacitance increases the spiking frequency, while decreasing the reset capacitance lowers the energy per spike. Unlike the LIF neuron, however, the ML neuron keeps its EPS relatively low over a broader range of $C_{res}$ values—staying below 25 aJ even when $C_{res}$ is as large as 7 fF. This implies that the ML circuit can accommodate larger capacitances without incurring a prohibitive energy penalty, partially due to its "direct-drive" mechanism mirroring the calcium influx in the original Morris–Lecar equations.

Across the parameter sweeps, the lowest EPS of 0.942 aJ emerges at $C_{mem}$ = 1 fF and $C_{res}$ = 0.1 fF, while the highest frequency (1.06 Ghz) is achieved at $C_{mem}$ = 0.4 fF and $C_{res}$ = 0.1 fF. Balancing these extremes, Figure 4(c) reveals that a 0.4 fF–0.1 fF configuration provides an optimal trade-off between frequency and EPS—directly mirroring the multi-modal firing regime predicted by Morris–Lecar equations when key parameters (here, capacitances) are well-tuned.

Further supply-voltage sweeps in Figures 4(d–f) underscore the neuron's subthreshold operation strengths: with $C_{mem}$ = 0.4 fF and $C_{res}$ = 0.11 fF, the ML neuron maintains a maximum value of 954 MHz firing rate at $V_{supp}$ = 0.25V. This high-performance region extends up to 0.32 V, which corresponds to very low energy per spike below 1 aJ. This behavior reflects the bifurcation properties of the ML model—in deep subthreshold, the device gracefully transitions between low-energy bursting and stable spiking modes[42]. The minimum EPS is recorded at 0.385 aJ at 0.14 V, illustrating how the ML's biophysical grounding (i.e., voltage-gated channel emulation) can support highly efficient spike generation. A noisy supply voltage could significantly affect the performance of other neuron models and their respective configurations[43–45]. However, from a practical standpoint, these results confirm that the ML neuron—by virtue of its direct membrane drive—can tolerate moderate supply-voltage fluctuations (0.1 V–0.3 V) with little degradation in performance. After normalizing the data, the optimization score shows that $V_{supp}$ = 0.25V produces the most optimal balance between energy per spike and spiking frequency (Figure 4f), yielding 1.01 aJ/spike at 954 MHz—a notably efficient operating point. Compared to the LIF neuron's decoupled mirror approach, this ML configuration offers a better balance of low energy and stable high-frequency spiking, making it especially attractive for noisy or resource-constrained environments.

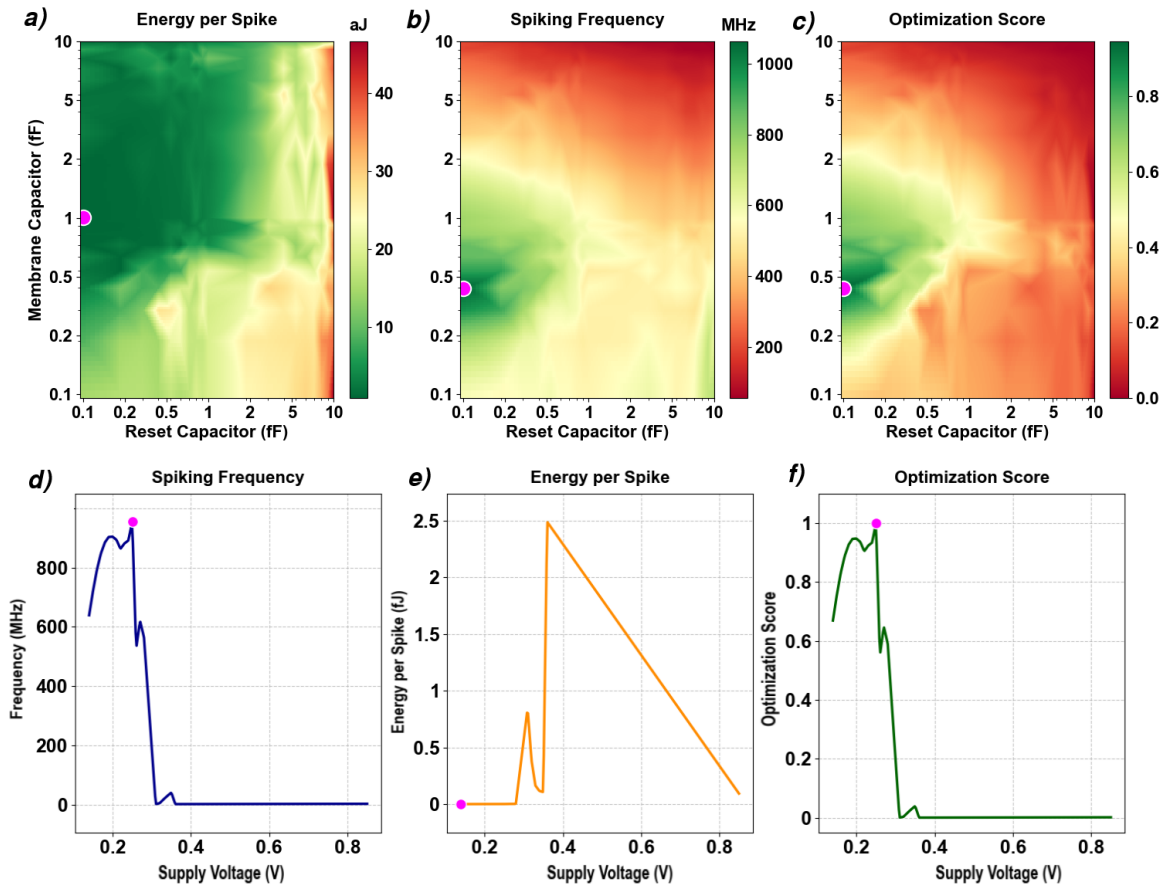

Figure 4. Parametric analysis of the Morris–Lecar (ML) neuron in 7 nm FinFET technology. (a–b) Contour plots showing how membrane (vertical axis) and reset (horizontal axis) capacitances affect (a) energy per spike (aJ) and (b) spiking frequency (MHz), averaged over supply voltages from 0.1 V to 0.6 V. The minimal observed energy (~0.942 aJ) and maximal frequency (~1.06 GHz) occur in regions of lower capacitance. (c) The combined optimization score pinpoints 0.4 fF–0.1 fF (membrane–reset) as the best-performing configuration, balancing both efficiency and speed. (d–e) With that configuration, sweeping Vsupp from 0.1 V to 0.5 V reveals a maximum frequency of ~954 MHz near 0.3 V and a minimum energy of 0.385 aJ at 0.14 V. (f) The optimization score across the same voltage sweep identifies 0.25 V as the desired spot, yielding ~1.01 aJ/spike at 954 MHz and demonstrating the ML neuron's resilience in deep-subthreshold operation.

**AH Neuron Optimization**

Recall from the Axon-Hillock Model subsection that the AH neuron merges the membrane and reset nodes into a single capacitor $C_{res}$. Figure 5a-c is now a line plot rather than a contour, illustrating how the single capacitance influences spiking frequency and energy per spike. Strikingly, smaller capacitances yield both higher frequency and lower energy consumption— a direct contrast to the LIF and ML neurons, where decreasing one capacitor typically improved frequency but increased energy or vice versa. In both the LIF and ML, a decreasing membrane capacitance generally sacrificed efficiency for higher spiking frequencies. However, in this neuron, the opposite is true. As our associated capacitors decrease in size, the dual modality of serving as both the integrator and reset mechanism gives way to increasing efficiency and spiking frequency simultaneously. Combined with the superior electrostatic control of the 7 nm FinFET devices, the AH neuron can achieve extremely rapid firing. The maximum frequency of 1.57 GHz is measured at $C_{res}$ = 0.12 fF and the minimum EPS of 12.48 aJ at $C_{res}$ = 0.25 fF. Despite the relatively higher EPS, it should be noted that EPS of AH neurons will significantly improve after sweeping and identifying the optimal supply voltage.

Figure 5(c) indicates that the optimal all-around capacitance setting for the AH neuron—averaged across multiple supply voltages—is Cres = 0.12 fF, producing a 1.57 GHz firing rate with an EPS of 19.2 aJ. Further voltage sweeps in Figure 5(d–f) confirm that lowering the supply voltage can dramatically cut the energy per spike (down to 0.252 aJ), at the expense of maximal frequency. Notably, the neuron maintains a high spiking rate around 2.5 GHz across a broad operating range (0.3 V to 0.6 V), showcasing strong resilience to supply-voltage biases[45]. This robust, high-frequency operation reflects the Hopf-like bifurcation properties of the AH model,[42] wherein the combined membrane–reset capacitor leads to finite-frequency oscillations near the threshold region. AH neuron shows 6 times higher firing frequencies than the ML neuron due to reducing transistor overhead and redundant charging paths by simplifying integration and reset mechanisms onto one capacitor. At the optimal supply voltage of 0.61 V (Figure 5f), the AH neuron fires at 3.05 GHz with an energy of 8.78 aJ per spike, substantially outperforming the LIF and ML neurons in raw throughput. Despite a somewhat higher baseline EPS compared to deep-subthreshold ML configurations, the AH design excels at ultra-fast spiking once the supply voltage is near or above the device threshold. Overall, these results confirm that the single-capacitor Axon-Hillock topology—closely mirroring the biological axon hillock's all-or-nothing firing—offers a compelling blend of high speed, low energy, and stable operation under near-threshold conditions.

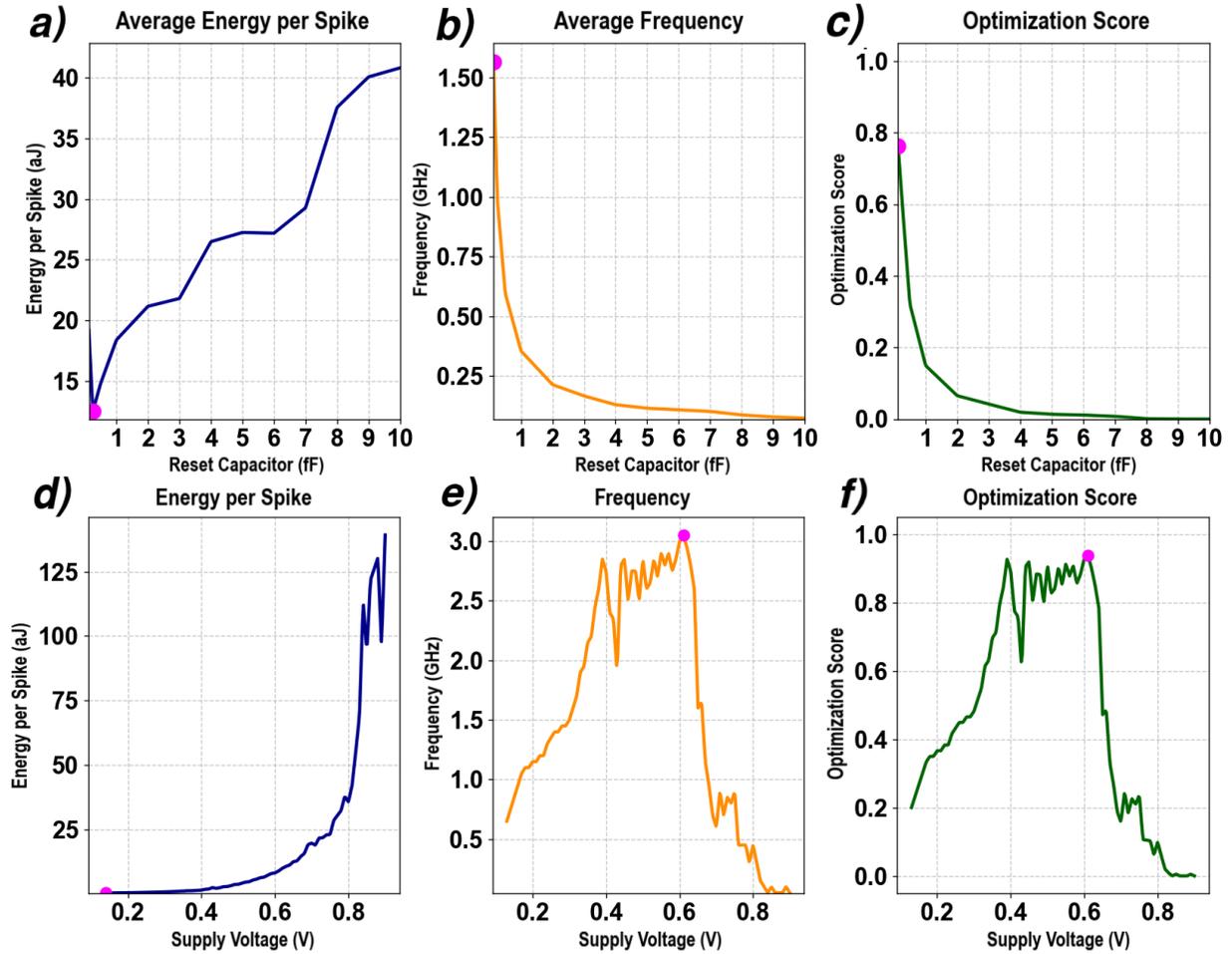

Figure 5. Parametric analysis of the Axon-Hillock (AH) neuron in 7 nm FinFET technology. (a–c) Plots of (a) energy per spike (aJ), (b) spiking frequency (GHz), and (c) the combined optimization score versus reset capacitance, averaged over supply voltages from 0.1 V to 0.9 V. The minimum average energy (~12.48 aJ) and maximum average frequency (~1.57 GHz) occur at small reset capacitances, with the optimal point at ~0.12 fF. (d–f) Using the best-performing reset capacitor from (a–c), a voltage sweep further illustrates how the AH neuron behaves under different Vsupp settings. (d) Energy per spike can drop as low as 0.252 aJ at low Vsupp, (e) spiking frequency peaks at 3.05 GHz near 0.61 V, and (f) the optimization score confirms 0.61 V as the sweet spot balancing ultra-high frequency and low energy.

Lastly, we analyze the static power consumption of each neuron—i.e., the power drawn when synaptic input is negligible and the circuit remains near its resting state. Figure 6 depicts static power scales with supply voltage for the LIF, ML, and AH neurons. Notably, AH shows a relatively modest, nearly linear rise as $V_{supp}$ increases from subthreshold levels to 0.9 V, reflecting the single-capacitor design's minimal leakage pathways. In contrast, both LIF and ML exhibit a steeper exponential trend past 0.6 V, dissipating an order of magnitude more static power than AH at that voltage. This difference aligns with the previous parametric findings—namely, that the ML and LIF circuits each introduce additional transistors for "mirroring" or "leak" paths, which become more prominent at higher supply voltages. By comparison, the AH neuron's simpler design places fewer devices in a leakage-prone state, thereby preserving lower static power. These observations highlight how circuit topology, particularly the number and arrangement of transistors responsible for membrane integration and reset, substantially impacts low-voltage leakage currents in advanced 7 nm FinFET technology.

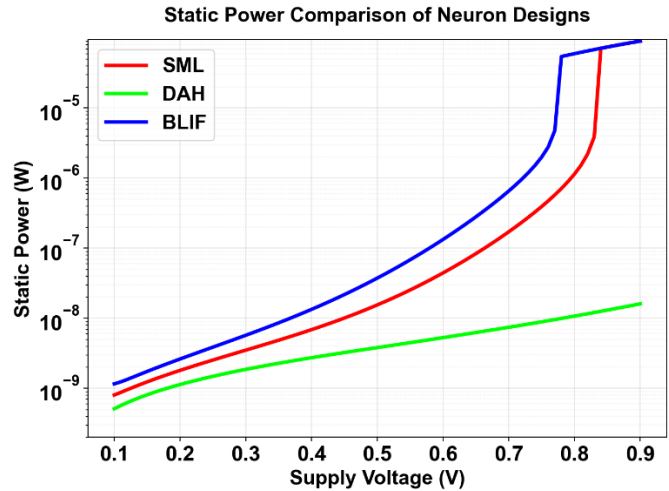

Figure 6. Static power consumption versus supply voltage for the three 7 nm FinFET neuron designs (LIF, ML, AH).

## Discussion

Table 1 summarizes the performance of three neuron models—LIF, ML, and AH—across multiple CMOS nodes, including 22 nm planar (from our prior work) and the 7 nm FinFET designs presented here. Several trends emerge:

**Subthreshold Leakage vs. High-Speed Operation**

At supply voltages below ~0.3 V, the 22 nm planar circuits exhibit lower static leakage than our 7 nm FinFET designs. This contrast reflects the short-channel effects in aggressively scaled FinFETs, where higher doping densities and smaller gate lengths elevate subthreshold conduction[46]. However, once the supply voltage surpasses approximately 0.3 V, the FinFET circuits deliver markedly higher spiking frequencies and more efficient dynamic switching. For instance, LIF in 7 nm achieves a 1.66 GHz firing rate, whereas the equivalent 22 nm LIF circuit peaks at 343 kHz. Thus, the improved electrostatic control in FinFETs outweighs the leakage penalty at moderately higher voltages.

**Comparative Merits of LIF, ML, and AH**

Leaky-Integrate-and-Fire decouples output from input, enabling high firing rates but with relatively higher energy overhead—1.26 aJ/spike at its best-operating point in 7 nm. Despite scaling improvements, LIF still sees steeper static-power growth once $V_{supp}$ exceeds 0.6 V (Figure 6) due to added transistors for current mirroring. Morris–Lecar trades absolute speed for better efficiency (0.942 aJ/spike) thanks to its direct-drive membrane approach, which omits the LIF's current mirror. It supports bursting behavior, further reducing average power in subthreshold regimes (up to ~0.3 V). Although peak firing frequency (1.06 GHz) trails the LIF design (1.66 Ghz), the ML neuron remains robust under noisy or resource-constrained settings. Axon-Hillock capitalizes most on FinFET characteristics, merging membrane and reset capacitance into a single capacitor. This simplification minimizes leakage and parasitic pathways, leading to multi-gigahertz firing rates (3.05 GHz) with sub-10 aJ energies near 0.6 V. While subthreshold leakage at 0.1 V is higher than 22 nm[47] ($3.0 \times 10^{-9}$ W vs. $1.0 \times 10^{-11}$ W), the AH neuron greatly outperforms the planar node at moderate-to-high voltages, reinforcing why advanced nodes are ideal for high-throughput spiking circuits.

**Trade-Offs in Capacitor Sizing**

Across all neurons, decreasing either the membrane or reset capacitor raises the spiking frequency (due to faster charging), but can also affect energy. A carefully tuned reset capacitor greatly reduces the energy per spike by shrinking the discharge overhead. This balance appears clearly in the AH design, where the single-capacitor approach simultaneously improves frequency and efficiency at small capacitances.

In summary, each neuron architecture—LIF, ML, AH—has distinctive strengths (e.g., highest speed, best efficiency, or burst behavior). Migrating these designs from 22 nm planar to 7 nm FinFET greatly enhances peak spiking frequency and lowers dynamic power at moderate supply voltages, despite somewhat higher leakage in deep subthreshold. Consequently, the AH neuron in 7 nm achieves the best overall performance among the three models, underscoring that advanced FinFET nodes can dramatically improve neuromorphic circuits' speed–efficiency trade-offs.

| Neuron Model | Leaky-Integrate-and-Fire | | | Morris-Lecar | | | Axon-Hillock | | |
|---|---|---|---|---|---|---|---|---|---|
| Tech | 28 nm [27] | 22 nm [47] | 7 nm * | 65 nm [33] | 22 nm [47] | 7 nm * | 65 nm [34] | 22 nm [47] | 7 nm * |
| Optimal $V_{supp}$ | 0.25 V | 0.2 V | 0.47 V | 0.175 V | 0.7 V | 0.21 V | 0.2 V | 0.4 V | 0.61 V |
| Design | Analog | Analog | Analog | Analog | Analog | Analog | Analog | Analog | Analog |
| Scale | Acc. | Acc. | Acc. | Bio. | Bio. | Bio. | Bio. | Bio. | Bio. |
| $I_{syn}$ | 1.5 nA | 100 nA | 100 nA | 50-150 pA | 100 nA | 100 nA | 150 pA | 100 nA | 100 nA |
| $C_{mem}$ | 3.47 fF | 1 fF | 0.5 fF | 4 fF | 1 fF | 0.4 fF | 5 fF | 1 fF | 0.12 fF |
| $EPS_{min}$ | 1.61 fJ | 300.4 aJ | 2.35 aJ | 4 fJ | 4 fJ | 0.942 aJ | 2 fJ | 241.1 aJ | 8.78 aJ |
| $F_{max}$ | 300 kHz | 1.37 MHz | 1.66 Ghz | 26 kHz | 390 Mhz | 1.06 GHz | 15.6 kHz | 80 Mhz | 3.05 GHz |
| $P_{static}$ (~0.1V) | - | $2e^{-10}$ W | $1e^{-9}$ W | $1e^{-10}$ W | $2e^{-10}$ W | $2e^{-9}$ W | $3e^{-11}$ | $1e^{-11}$ W | $3e^{-9}$ W |
| $P_{static}$ (~0.9V) | - | $3e^{-4}$ W | $3e^{-4}$ W | - | $3e^{-4}$ W | $3e^{-4}$ W | - | $8e^{-7}$ W | $9e^{-7}$ W |
| Area | 34 $\mu m^2$ | - | - | 35 $\mu m^2$ | - | - | 31 $um^2$ | - | - |

Table 1. Comparison of 7 nm FinFET Configuration/Performance to Original Works. *Shows the results of this work

## Methods

### Simulation Environment

All simulations were conducted using the ASAP 7 nm FinFET predictive PDK (ASAP7)[25,26] with the BSIM-CMG3[48] compact model in NGSpice. The ASAP7 models assume a single fin for both PMOS and NMOS in truly complementary configurations, ensuring the same number of PMOS and NMOS fins in each branch. In a standard inverter, reducing the fin count for PMOS relative to NMOS would lower fan-out, reflecting the model's greater mobility in PMOS devices. For this work, we maintained one fin per device to balance performance and simplicity.

NGSpice's GMIN convergence algorithm was configured with up to 1000 iterations and a minimum step size small enough to resolve rapid spiking transitions. These settings provided stable numerical solutions while retaining accuracy for subthreshold and near-threshold operation.

### Measuring Key Performance Metrics

**Spiking Frequency ($f_{spike}$):** We identified each neuron's periodic or bursting output by detecting rapid voltage transitions above a threshold. The time between successive spikes yields the firing frequency in Hertz (or MHz/GHz).

**Energy per Spike ($E_{spike}$):** We computed Espike by integrating instantaneous power dissipation in every transistor and capacitor during one spiking cycle. This integral captures both dynamic switching events (charging/discharging) and any static leakage over the spike period.

**Static Power ($P_{static}$):** To measure idle (static) power consumption, we minimized synaptic input current (near zero) and observed the power dissipation at equilibrium. Taking an average of a steady-state window helps eliminate transient effects, isolating baseline leakage.

### Parametric Sweeps

We systematically varied:

**Supply voltage ($V_{supp}$):** 0.1 V to 0.9 V in 0.05 V increments, spanning deep subthreshold to near-nominal conditions.

**Membrane ($C_{mem}$) and reset ($C_{res}$) capacitances:** 0.1 fF to 10 fF in integer or half-integer steps, covering very small values favored in modern neuromorphic designs as well as larger sizes for backward comparison with older nodes (e.g., 180 nm, 65 nm).

For each (Vsupp, Cmem, Cres) combination, we recorded fspike and Espike (and static power, when the neuron is idle). We then normalized these values to highlight optimal operating points that trade off spiking frequency and energy. Specifically, the aggregated optimization score balanced high frequency with low energy, as shown in Equations (7)–(9).

## Data Availability

The code and data generated or analyzed during this study are available in the GitHub repository:
https://github.com/INQUIRELAB/7nm-FinFet-Spiking-Neuron-Simulations.git

## Author contributions statement

## Competing Interest Statement

The authors declare that they have no competing financial or non-financial interests that could have influenced the work reported in this paper.